\documentclass{article}
\usepackage{spconf,amsmath,graphicx}
\usepackage{amssymb}
\usepackage{booktabs}
\usepackage{pifont}
\usepackage{multirow}
\usepackage{cancel}
\usepackage{tikz}
\usetikzlibrary{positioning}
\usetikzlibrary{calc}
\usepackage{pgfplots, pgfplotstable}    
\pgfplotsset{compat=1.16} 
\usepackage{url}
\usepackage{hyperref}
\usepackage{algorithm,algpseudocode}
\DeclareMathSymbol{\shortminus}{\mathbin}{AMSa}{"39}

\newcommand{\Sref}[1]{\S\ref{#1}}
\newcommand{\Fref}[1]{Figure~\ref{#1}}
\newcommand{\cmark}{\ding{51}}%
\newcommand{\Tref}[1]{Table~\ref{#1}}

\newcommand{\Hquad}{\hspace{0.0em}} 
\newcommand\mypar[1]{\noindent\textbf{#1}\Hquad}

\usepackage{icomma}

\makeatletter
\def\blfootnote{\xdef\@thefnmark{}\@footnotetext}
\makeatother

\usepackage[
backend=biber,
style=ieee,
citestyle=numeric-comp,
maxbibnames=3,
maxcitenames=3,
doi=false,isbn=false,url=false,eprint=false
]{biblatex}

\defbibheading{bibliography}[\refname]{}

\title{Cross-Modal Multi-Tasking for Speech-to-Text Translation\\via Hard Parameter Sharing}

\name{
\begin{tabular}{c}
\it Brian Yan${}^{1}$, Xuankai Chang${}^1$, Antonios Anastasopoulos${}^3$, Yuya Fujita${}^4$, Shinji Watanabe${}^{1,2}$
\end{tabular}
}
\address{${}^1$Carnegie Mellon University, US, ${}^2$Johns Hopkins University, US,\\${}^3$George Mason University, US, ${}^4$Yahoo Japan Corporation, JP}

\begin{document}
\ninept
\maketitle
\begin{abstract}
Recent works in end-to-end speech-to-text translation (ST) have proposed multi-tasking methods with \textit{soft parameter sharing} which leverage machine translation (MT) data via secondary encoders that map text inputs to an eventual cross-modal representation.
In this work, we instead propose a ST/MT multi-tasking framework with \textit{hard parameter sharing} in which all model parameters are shared cross-modally.
Our method reduces the speech-text modality gap via a pre-processing stage which converts speech and text inputs into two discrete token sequences of similar length -- this allows models to indiscriminately process both modalities simply using a joint vocabulary.
With experiments on MuST-C, we demonstrate that our multi-tasking framework improves attentional encoder-decoder, Connectionist Temporal Classification (CTC), transducer, and joint CTC/attention models by an average of +0.5 BLEU without any external MT data.
Further, we show that this framework incorporates external MT data, yielding +0.8 BLEU, and also improves transfer learning from pre-trained textual models, yielding +1.8 BLEU.\footnote{Recipes and models are available in \href{https://github.com/espnet/espnet/tree/master/egs2/must\_c\_v2/st2}{ESPnet (egs2/must\_c\_v2/st2)}.}
\end{abstract}

\section{Introduction}

\label{sec:intro}

One of the preeminent challenges in end-to-end speech-to-text translation (ST) is that of data scarcity. 
There are relatively small amounts of labeled ST data compared to automatic speech recognition (ASR) and machine translation (MT) data, as well as unpaired speech and text data.
Simply pseudo-labeling ASR data with strong MT models has proven to be effective \cite{pino2019harnessing, inaguma2021source}; however, synthesizing speech for MT data using TTS has proven to be more complex and less effective \cite{pino2019harnessing, jia2022leveraging}.
So what techniques aside from data augmentation can leverage textual data towards improving ST systems?

A popular answer amongst recent works is multi-tasked learning, where models are jointly optimized to perform MT and ST.
Many proposed multi-tasking methods employ varying degrees of \textit{soft parameter sharing} \cite{ruder2017overview}, where some parameters are shared while others are task-specific.
Generally, these methods use modality-specific modules that map continuous speech inputs and discrete text inputs into an eventual common latent space \cite{ao2021speecht5, bapna2022mslam, tang2022unified, ye2022cross, fang-etal-2022-stemm, chen2022maestro, cheng2023m, ouyang-etal-2023-waco, wu2023decoder}.
In this work, we refer to this family of approaches as \textbf{soft multi-tasking} methods.
These methods often require cross-modal regularization \cite{ye2022cross, fang-etal-2022-stemm, chen2022maestro, cheng2023m, ouyang-etal-2023-waco} to encourage greater similarity between speech and text representations, demonstrating that the gap speech and text modalities must be reduced to enable the benefits of soft multi-tasked learning.

\begin{figure}
\centering
\includegraphics[width=0.9\linewidth]{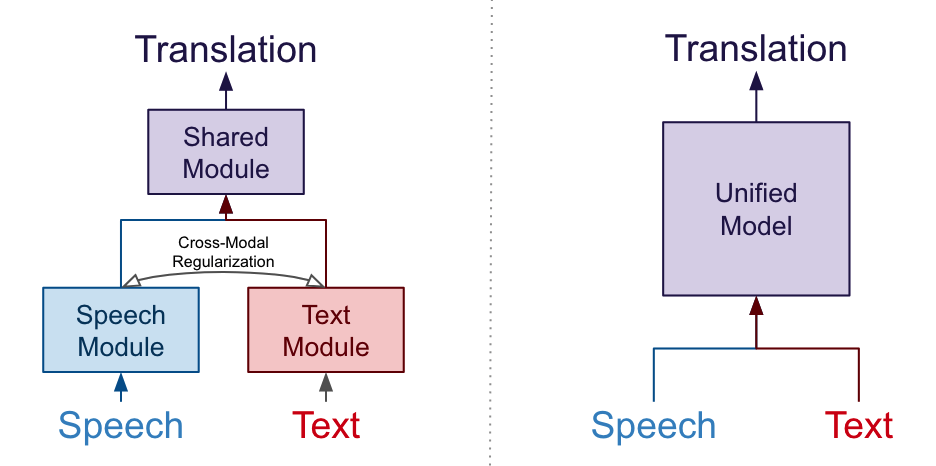}
\vspace{-2mm}
\caption{Illustrative examples of soft (left) vs. hard (right) parameter sharing approaches to ST/MT multi-tasking.}
\label{fig:illustrative}
\vspace{-2mm}
\end{figure}

An alternative approach to reducing the speech-text modality gap is to first convert continuous speech signals into discrete sequences.
Recent works have shown that discrete speech sequences obtained by applying $k$-means clustering on self-supervised learning (SSL) representations contain sufficient semantic information to be effective system inputs for ASR \cite{chang23b_interspeech} and ST \cite{rubenstein2023audiopalm}. Similarly, speech discretization has been applied in TTS \cite{hayashi2020discretalk, lakhotia2021generative, wang2023neural, hassid2023textually} and speech-to-speech translation (S2ST) \cite{lee2022direct, li2023textless, rubenstein2023audiopalm}.
These developments in speech discretization appear to drastically reduce the speech-text modality gap, suggesting that the modality-specific modules employed by soft parameter sharing methods may not be necessary.

In this work we investigate \textit{hard parameter sharing} \cite{ruder2017overview}, where a single unified architecture handles both ST and MT without any modality-specific modules or cross-modal regularization -- we posit that this \textbf{hard multi-tasking} approach can:
\begin{enumerate}
    \item Be generally applicable to any sequence-to-sequence model
    \item Be used to incorporate external MT data
    \item Improve transfer learning from pre-trained textual models 
\end{enumerate}

Inspired by the recent AudioPalm work \cite{rubenstein2023audiopalm} which demonstrates hard ST/MT multi-tasking via a decoder-only model,\footnote{Please refer to \Sref{sec:prior-work} for a full accounting of the novelties in this work.} we investigate a similar concept for attentional encoder-decoder (AED), Connectionist Temporal Classification (CTC), transducer (RNN-T), and joint CTC/attention (CTC/Attn) models.
Our method pre-processes speech and text inputs to produce two discrete sequences of comparable length; \textbf{doing so allows ST/MT multi-tasking to be realized through a single token-to-token sequence model}.
Our models thus ingest speech and text inputs simply using a joint vocabulary.
Intuitively, the speech modality is treated as another ``language'' represented by a distinct writing system.
Specifically, speech is \textit{discretized} via $k$-means clustering over SSL representations (e.g. WavLM \cite{chen2022wavlm}) and \textit{down-sampled} via repetition removal and subword tokenization while text is \textit{up-sampled} via token repetition.

In our experiments, we first show that our hard parameter sharing approach improves AED, CTC, RNN-T, and CTC/Attn models when trained from scratch by an average of +0.5 BLEU on the MuST-C ST corpus \cite{di2019must} without any external MT data (\Sref{sec:res_1}).
Next, we show that leveraging external WMT \cite{bojar2016findings} MT data via our multi-tasking framework yields an additional +0.8 BLEU (\Sref{sec:res_2}).
Finally, we show that our multi-tasking approach also improves the efficacy of transfer learning by +1.8 BLEU from pre-trained textual models (e.g. mBART \cite{liu2020multilingual, tang2020multilingual}) (\Sref{sec:res_3}).

\section{Proposed Framework}
\label{sec:proposed}

In this section, we first describe our method of converting continuous speech into sequences of discrete units (\Sref{sec:discretization}) and then explain how discrete speech inputs enable \textbf{hard parameter sharing} (\Sref{sec:hard-sharing}).
\Sref{sec:seq2seq} describes the set of sequence-to-sequence (seq2seq) frameworks which we investigate.

\begin{figure}
\centering
\includegraphics[width=\linewidth]{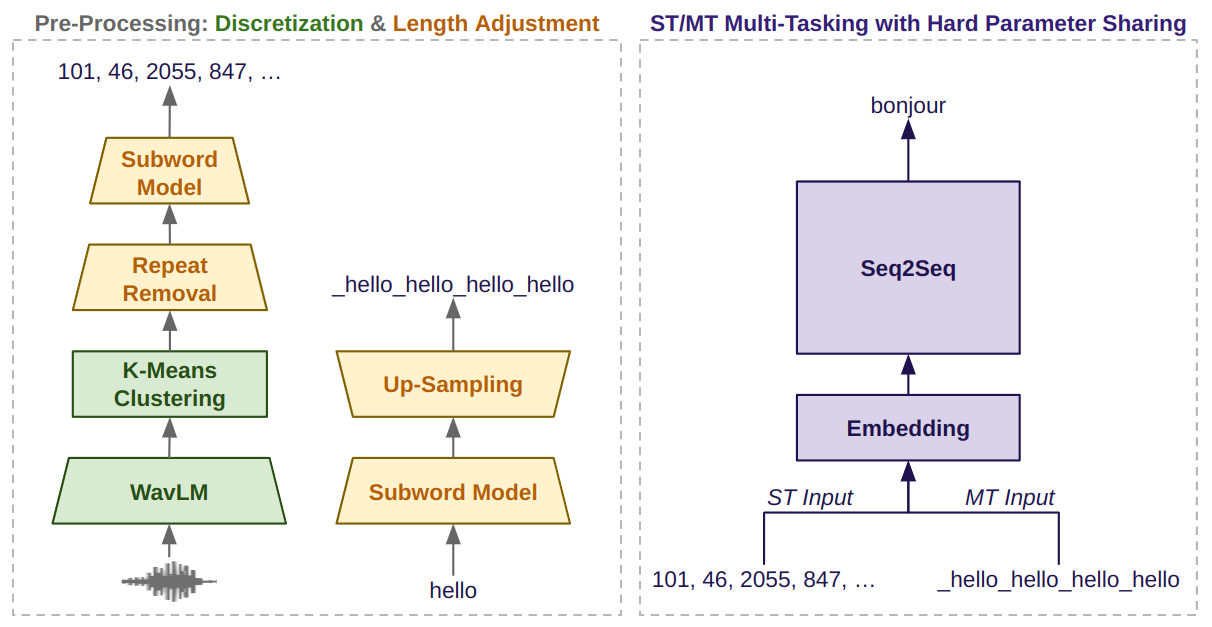}
\vspace{-2mm}
\caption{Speech/text pre-processing (left) produces two discrete sequences of similar length which are ingested by Seq2Seq models (right) with hard parameter sharing between ST/MT tasks.
}
\label{fig:main}
\end{figure}

\Fref{fig:main} also summarizes our approach. The color-coding in the figure matches bolded keywords in the following section.

\subsection{Speech Discretization}
\label{sec:discretization}

The objective of speech discretization is to convert a continuous speech signal, $X^{\textsc{cont}} = \{ \mathbf{x}_t \in \mathbb{R}^D | t = 1, ..., T \}$, into a discrete sequence, $X^{\textsc{disc}} = \{ x_l \in \mathcal{V}^{\textsc{spe}} | l = 1, ..., L \}$ where $\mathcal{V}^{\textsc{spe}}$ is a vocabulary consisting of some discrete units representing chunks of speech.
There are many discretization techniques which can accomplish this \cite{van2017neural,zeghidour2021soundstream,wang2023neural,defossez2022high}; we opt for the approach described by \cite{chang23b_interspeech} which uses $k$-means clustering and SSL representations in a manner similar to HuBERT \cite{hsu2021hubert}.

We first select an appropriate SSL model and intermediate layer to work with.
In this work, we use \textcolor[HTML]{1E8449}{\textbf{WavLM}} which is pre-trained on 94k hours of English speech with masked prediction and de-noising objectives \cite{chen2022wavlm}.
We select the 21st layer from WavLM as it has the highest canonical correlation analysis (CCA) similarity to word labels \cite{pasad2023comparative}, suggesting that these representations contain useful semantic information.
Next, we extract these representations for a portion of our training data and use them to train a $k$-means model with $2000$ centroids.
We then use this $k$-means model to convert the entire training set into sequences of \textcolor[HTML]{1E8449}{\textbf{$k$-means cluster}} assignments; note that at this point we have a sequence of discrete units but have only down-sampled the sequence length to 50 kHz via WavLM.

As noted by prior works \cite{lee2022direct, chang23b_interspeech}, these sequences of $k$-means cluster assignments can be collapsed by \textcolor[HTML]{CA6F1E}{\textbf{removing repeats}} of the same consecutive unit.
Finally, \textcolor[HTML]{CA6F1E}{\textbf{subword modeling}} can be applied to further reduce the sequence length \cite{chang23b_interspeech}; we use the unigram algorithm from SentencePiece \cite{kudo2018sentencepiece} to construct a vocabulary of $4000$ tokens.
To reduce over-fitting on particular segmentation patterns, we also apply BPE-dropout \cite{provilkov2020bpe} during training.
Ultimately, we obtain discrete speech sequences with an average length of 122 tokens from original audio with an average duration of 6.4 seconds.

\subsection{ST/MT Multi-Tasking with Hard Parameter Sharing}
\label{sec:hard-sharing}

\subsubsection{ST from Discrete Speech Inputs}

With discrete speech inputs, the ST task seeks to map a sequence of tokens $X^{\textsc{disc}}$ into another sequence of tokens $Y^{\textsc{tgt}} = \{ y^{\textsc{tgt}}_m \in \mathcal{V}^{\textsc{tgt}} | l = 1, ..., M \}$ where $\mathcal{V}^{\textsc{tgt}}$ is the subword vocabulary for the target language -- this is analogous to the MT task.
We therefore replace the convolutional feature extractor typically used by systems which process continuous speech inputs \cite{prabhavalkar2023end} with a learned \textcolor[HTML]{76448A}{\textbf{embedding}} layer \cite{stahlberg2020neural}.
Following \cite{chang23b_interspeech}, we apply time-masking to the sequences produced by the embedding layer; this is an additional form of data augmentation similar to SpecAugment \cite{park2019specaugment} which is commonly applied to continuous speech inputs.

Our ST data is defined as a set of speech, source text, and target text triplets $\{(X^{\textsc{disc}}, Y^{\textsc{src}}, Y^{\textsc{tgt}})\}$. Similar to target text, source text is a sequence of tokens $Y^{\textsc{src}} = \{ y^{\textsc{src}}_n \in \mathcal{V}^{\textsc{src}} | l = 1, ..., N \}$ where $\mathcal{V}^{\textsc{src}}$ is the subword vocabulary for the source language.

\subsubsection{Incorporating the MT Task}
\label{sec:mt}

Since the discrete speech sequences $X^{\textsc{disc}}$ are still longer than their corresponding source text $Y^{\textsc{src}}$, we repeat the source text tokens by a factor of $4$ to further reduce the gap between speech and text inputs (see \Sref{sec:res_4} for \textcolor[HTML]{CA6F1E}{\textbf{up-sampling}} factor ablations).
For instance, the sequence ``\_a\_b'' becomes ``\_a\_a\_a\_a\_b\_b\_b\_b''.
This length adjustment approach has been shown to be effective for injecting text into speech models \cite{renduchintala2018multi, thomas2022}.
We denote this up-sampled source text as $X^{\textsc{text}}$ and define our MT data as triplets of up-sampled source text, source text, and target text $\{(X^{\textsc{text}}, Y^{\textsc{src}}, Y^{\textsc{tgt}})\}$.

To incorporate textual inputs into our discrete ST models, we simply extend the input vocabulary $\mathcal{V}^{\textsc{cross}} = \mathcal{V}^{\textsc{spe}} \cup \mathcal{V}^{\textsc{src}}$ to include textual subword tokens $\mathcal{V}^{\textsc{src}}$ from the source language in addition to the speech subword tokens $\mathcal{V}^{\textsc{spe}}$.
This modification correspondingly expands the \textcolor[HTML]{76448A}{\textbf{embedding}} layer, but does not impact any other component in the architecture.
Note that this joint speech-text vocabulary allows our models to indiscriminately ingest speech or text into any \textcolor[HTML]{76448A}{\textbf{seq2seq}} model, sharing all non-embedding parameters between ST and MT.
The only modality-specific parameters are within the embedding layer, as speech and text tokens are still disjoint.

Now MT multi-tasking can be achieved by simply combining ST and MT training data: the training set consists of triplets  $\{(X^{\textsc{cross}}, Y^{\textsc{src}}, Y^{\textsc{tgt}})\}$ where $X^{\textsc{cross}} = \{ x_l \in \mathcal{V}^{\textsc{cross}} | l = 1, ..., L \}$.
Note that the source text $Y^{\textsc{src}}$ and the target text $Y^{\textsc{tgt}}$ are identical for ST and MT examples.
The same losses (described in the following section) are applied with equal weighting between the two tasks.
\textit{All} parameters are updated in each iteration.
Models do not have any explicit sense of whether a particular example is an MT or ST task -- all are processed in the same manner.

\subsection{Seq2Seq Models}
\label{sec:seq2seq}

In this work, we examine AED \cite{bahdanau_attn, chan2015listen}, CTC \cite{graves2006connectionist}, RNN-T \cite{rnnt_graves}, and CTC/Attn \cite{watanabe2017hybrid, yan2023ctc} models.
We use a hierarchical encoding scheme, as in \cite{yan2023ctc}, for all of our models.
This method applies an ASR CTC objective at an intermediate encoder layer, denoted as $\mathcal{L}_{\textsc{src\_ctc}}$, and a second ST CTC objective at the final encoder layer, denoted as $\mathcal{L}_{\textsc{tgt\_ctc}}$.
The ASR CTC objective allows our models to utilize source language transcriptions to improve encoder representations \cite{inaguma2020espnet, gaido2021ctc}.
The ST CTC objective acts as a form of regularization which encourages encoder representations to be monotonic with respect to the target sequence; this has been shown to improve the translation quality of auto-regressive systems \cite{zhang2022revisiting, yan2023ctc}.
Our AED and CTC/Attn models use an additional cross-entropy loss, denoted as $\mathcal{L}_{\textsc{CE}}$, while our RNN-T models use an additional RNN-T loss, denoted as $\mathcal{L}_{\textsc{RNNT}}$.
AED and RNN-T models are jointly trained with CTC losses but CTC likelihoods are not applied during decoding.
All told, our models are optimized using an interpolated loss defined as $\mathcal{L} = \lambda_1 \mathcal{L}_{\textsc{src\_ctc}} + \lambda_2 \mathcal{L}_{\textsc{tgt\_ctc}} + \lambda_3 \mathcal{L}_{\textsc{CE/RNNT}}$. 
We use $\lambda_1=\lambda_2=0.3$ and $\lambda_3=1$ for our experiments. 
For CTC models, the last term is omitted and we use $\lambda_1=\lambda_2=1$.

\section{Experimental Setup}
\label{sec:setup}

We compare the performance of our \textbf{hard multi-task models} vs. \textbf{single-task baselines} with identical architectures.
The single-task baselines are also discrete ST systems which allow us to understand the effects of ST/MT multi-tasking, holding all else equal.
Note that the purpose of this work is not to prove the efficacy of systems with discrete speech inputs compared to those with continuous spectral inputs; this aspect has been addressed elsewhere \cite{chang23b_interspeech}. 
We use the ESPnet-ST-v2 toolkit \cite{yan-etal-2023-espnet} for our experiments. 
\newline

\mypar{Data:}
We use the En-De, En-Es, and En-Fr portions of MuST-C \cite{di2019must} which consist of 408/504/492 audio hours and $234$K/$270$K/$292$K sentences.
Experiments using external MT data were conducted on the En-De language pair using the WMT'16 corpus \cite{bojar2016findings} which consists of $4.6$M sentences.
Speed perturbation is applied to up-sample ST data by a factor of $3$.

\mypar{Models:}
We use separate vocabularies of $4000$ subword units built from MuST-C data for discrete speech, source text, and target text.
Unified multi-tasking models use a combined speech-text vocabulary consisting of $8000$ units, obtained by combining discrete speech and source text vocabularies.
Models with mBART initializations adopt the pre-trained model's $250$K target vocabulary. 

All models use input embedding with $1024$ dim.
We use 18 layer E-Branchformer \cite{kim2023branchformer} encoders with ASR CTC applied on the 12th layer.
Our base size models (denoted by $\texttt{A-D}$ the following section) use $256$ dim size, $1024$ feed-forward dim, and $4$ heads. 
Our larger models (denoted by $\texttt{E-F}$ the following section) use $512$ dim size, $2048$ feed-forward dim, and $8$ heads. 
We use 6 layer Transformer decoders for AED and CTC/Attn models with $2048$ feed-forward dim and either $4$/$8$ heads for base/large models.
Finally, for models with mBART we initialize only the decoder while freezing feed-forward and self-attention parameters and use 2x convolutional down-sampling after the encoder, following \cite{li2021multilingual, yan2023cmu}.

To ensure fair comparison, all models are trained for the same number of iterations regardless the training data size.
All model converge within 350K iterations and we average the 10 best checkpoints.
AED, RNN-T, and CTC/Attn models use beam search with beam size 10.
CTC models use greedy decoding.

\mypar{Evaluation:} 
We measure detokenized case-sensitive BLEU \cite{post2018call}.

\section{Results and Discussion}
\label{sec:results}

The objective of this work is to study several related, yet still distinct, dynamics within unified ST/MT multi-tasking.
We examine the effects of 1) \textbf{hard parameter sharing} for a set of sequence models (\Sref{sec:res_1}), 2) leveraging \textbf{external MT data} (\Sref{sec:res_2}), and 3) \textbf{transfer learning} from pre-trained textual models (\Sref{sec:res_3}).
We also provide ablations over sequence lengths (\Sref{sec:res_4}).

\subsection{Hard Parameter Sharing}
\label{sec:res_1}

\begin{table}[t]
  \centering
    \caption{Comparison (\textsc{bleu} scores) of single-task vs. hard ST/MT multi-task approaches for CTC, RNN-T, AED, CTC/Attn~models.}
    \resizebox {\linewidth} {!} {
\begin{tabular}{ll|c|cccc}
\toprule
 & & & \multicolumn{4}{c}{\underline{\textsc{MuST-C}}} \\
\texttt{\#} & \textsc{Model} & \textsc{Size} & En-De & En-Es & En-Fr & avg \\
\midrule
\multicolumn{7}{c}{\textsc{CTC}} \\
\midrule
\texttt{A1} & Single-Task & 50M & 23.2 & 27.9 & 32.2 & 27.8 \\
\texttt{A2} & \textbf{Hard Multi-Task} & 50M & \textbf{23.4} & \textbf{28.4} & \textbf{33.6} & \textbf{28.5} \\
\midrule
\multicolumn{7}{c}{\textsc{RNN-T}} \\
\midrule
\texttt{B1} & Single-Task & 60M & 26.4 & 30.4 & 33.1 & 30.0 \\
\texttt{B2} & \textbf{Hard Multi-Task} & 60M & \textbf{26.7}& \textbf{31.0} & \textbf{33.8} & \textbf{30.5} \\
\midrule
\multicolumn{7}{c}{\textsc{AED}} \\
\midrule
\texttt{C1} & Single-Task & 60M & 27.4 & 32.8 & 37.4 & 32.5 \\
\texttt{C2} & \textbf{Hard Multi-Task} & 60M & \textbf{27.7} & \textbf{33.1} & \textbf{38.0} & \textbf{33.0} \\
\midrule
\multicolumn{7}{c}{\textsc{CTC/Attn}} \\
\midrule
\texttt{D1} & Single-Task & 60M & 28.6 & 33.0 & 38.7 & 33.4 \\
\texttt{D2} & \textbf{Hard Multi-Task} & 60M & \underline{\textbf{29.2}} & \underline{\textbf{33.2}} & \underline{\textbf{39.2}} & \underline{\textbf{33.9}} \\
\bottomrule
\end{tabular}
}
    \label{tab:main}
    \vspace{-2mm}
\end{table}

As an alternative to soft parameter sharing methods which manage the ``distance'' between speech and text representations \cite{ye2022cross, fang-etal-2022-stemm, chen2022maestro, cheng2023m, ouyang-etal-2023-waco}, we employ a hard parameter sharing approach which uses a single set of parameters to capture both tasks.
To examine the effect of hard parameter sharing, we train models \textit{from scratch} and \textit{without external MT data} in this section.
Per \Sref{sec:mt}, we create an MT example $(X^{\textsc{text}}, Y^{\textsc{src}}, Y^{\textsc{tgt}})$ from each ST example $(X^{\textsc{disc}}, Y^{\textsc{src}}, Y^{\textsc{tgt}})$ and combine all ST/MT examples to form a training set.

The results in \Tref{tab:main} compare our hard multi-tasking method compared to the single-task baseline for CTC, RNN-T, AED, and CTC/Attn models.
We observe consistent improvements in the range of +0.2 to +1.4 BLEU points across three different language pairs and for all model types. 
We attribute these improvements to primarily to the regularization effect of hard parameter sharing \cite{ruder2017overview}, as we do not explicitly tell the model how to relate corresponding text and speech inputs.
Note that we observed the same trend when using the hierarchical encoding scheme (described in \Sref{sec:seq2seq}) as we did without; however, since this scheme produced better translation quality for all models, we chose to only present those results.

\subsection{Leveraging External MT Data}
\label{sec:res_2}

\begin{table}[t]
  \centering
    \caption{
    Performance of hard ST/MT multi-task CTC/Attn models with external MT data or mBART initialization. Single-task CTC/Attn baselines and soft multi-task models from prior works are shown for comparison.
    ${\dag}$Uses WMT16 MT data (4.6M sentences). $^{\ddag}$Uses external MT and unpaired text data via mBART initialization.
    }
    \resizebox {\linewidth} {!} {
\begin{tabular}{ll|c|cc|c}
\toprule
 & & & \multicolumn{2}{c|}{\underline{\textsc{Ext Data}}} & \\
\texttt{\#} & \textsc{Model} & \textsc{Size} & MT & Text & En-De \\
\midrule
- & ConST'22 \cite{ye2022cross} & 150M &  - & - & 25.7 \\
- & M$^{3}$ST'23 \cite{cheng2023m} & - &  - & - & 26.4 \\
\texttt{E1} & Single-Task CTC/Attn & 190M &  - & - & 29.0 \\
\texttt{E2} & \textbf{Hard Multi-Task CTC/Attn} & 190M &  - & - & \textbf{29.3} \\
\midrule
- & ConST'22 \cite{ye2022cross} & 150M &  \cmark$^{\dag}$ & - & 28.3 \\
- & M$^{3}$ST'23 \cite{cheng2023m} & - &  \cmark$^{\dag}$ & - & 29.3 \\
\texttt{E2+} & \textbf{Hard Multi-Task CTC/Attn} & 190M &  \cmark$^{\dag}$ & - & \textbf{30.1} \\
\midrule
\texttt{F1} & Single-Task CTC/Attn & 740M &  \cmark$^{\ddag}$ & \cmark$^{\ddag}$ & 29.3 \\
\texttt{F2} & \textbf{Hard Multi-Task CTC/Attn} & 740M &  \cmark$^{\ddag}$ & \cmark$^{\ddag}$ & \underline{\textbf{31.1}} \\
\bottomrule
\end{tabular}
}
    \label{tab:main2}
    \vspace{-4mm}
\end{table}

The ability to add external MT data into the training mixture is a major benefit of ST/MT multi-tasking.
Our approach is simple: we \textit{simply concatenate data sources} and train on the combined set.
The first two horizontal partitions of \Tref{tab:main2} present results on English-to-German ST with 4.6M sentences of external MT data from WMT'16.
Comparing \texttt{E1} to \texttt{E2+}, we see the full effect of multi-tasking with external MT data: +1.1 BLEU.
A portion of this gain must be attributed to that of hard parameter sharing.
To understand the impact of the external MT data on its own, we compare multi-tasking without external data, \texttt{E2}, to multi-tasking with external data, \texttt{E2+}: +0.8 BLEU.
Note that we can pre-train our entire model on this same external MT data, but this constitutes a form of transfer learning which we will discuss in the subsequent section.

We take two representative works for comparison in this section.
The ConST model \cite{ye2022cross} is a soft multi-tasking approach which utilizes a contrastive loss to encourage matched speech and text inputs to be closer, relative to unmatched speech and text inputs.
ConST also uses multiple strategies to create harder examples for the contrastive loss.
The M$^3$ST model \cite{cheng2023m} is another soft multi-tasking approach which utilizes a multi-stage training strategy.
The first stage is a purely textual pre-training stage which incorporates external MT data while the next two stages are ST fine-tuning stages which perform data mix-up and contrastive learning.
ConST and M$^3$ST appear to gain more from the same external MT data, although their baselines are indeed much weaker.
Nonetheless, we suspect that modality-specific modules can limit interference from extremely unbalanced MT to ST data ratios, but we leave this for future work.

\subsection{Cross-Modal Transfer Learning}
\label{sec:res_3}

\begin{table}[t]
  \centering
    \caption{Ablation study on the importance of up-sampling MT input text. MT = Use of MT multi-task. Up = Up-sampling factor of MT input text. Ratio = Average length ratio of discrete speech to text.}
    \begin{tabular}{cccc}
\toprule
MT & Up & Ratio & BLEU \\
\midrule
- & - & - & 28.6 \\
\cmark & - & 6.0 & 28.5 \\
\cmark & 2x & 3.0 & 28.8 \\
\cmark & 4x & 1.5 & \textbf{29.2} \\
\cmark & 6x & 1.0 & 28.7 \\
\bottomrule
\end{tabular}
    \label{tab:upsampling}
    \vspace{-4mm}
\end{table}

Ultimately, we'd like to build ST models which \textit{efficiently} leverage not only paired textual data, but also copious amounts of \textit{unpaired textual data}.
In this section we examine models initialized from mBART \cite{liu2020multilingual, tang2020multilingual}, an encoder-decoder pre-trained with text denoising objectives and then fine-tuned on large-scale MT data.\footnote{https://huggingface.co/facebook/mbart-large-50-many-to-many-mmt}
The recent trend is to take the mBART decoder parameters to partially initialize ST encoder-decoder models \cite{li2021multilingual, pham2022effective, zhang2022yitrans, yan2023cmu}.
This is a form of \textit{heterogeneous} transfer learning \cite{day2017survey} -- there is a cost associated with the distributional shift between the textual pre-training domain and the speech-based fine-tuning domain.
Cross-modal pre-training \cite{bapna2022mslam, tang2022unified} has been shown to reduce this cost. 
We posit that our cross-modal fine-tuning method has a similar effect.

The results in the final horizontal partition of \Tref{tab:main2} presents models with mBART decoder initialization (see \Sref{sec:setup}).
Comparing the single-task model \texttt{F1} with the multi-task model \texttt{F2}, we see that the latter is +1.8 BLEU better.
The single-task model only improves by +0.3 BLEU from mBART initialization (\texttt{E1} vs. \texttt{F1}); prior works have also noted similarly muted gains \cite{inaguma-etal-2023-unity, yan2023cmu}, indicating deficiencies in transfer learning across modalities.
Our method exhibits a more efficient transfer (\texttt{E2} vs. \texttt{F2}), yielding +1.8 BLEU.
Note that mBART has been fine-tuned on large scale MT data, so we do not find it necessary to include WMT data in our training mixture.

\vspace{-2mm}
\subsection{Ablations on Sequence Lengths}
\label{sec:res_4}

\Tref{tab:upsampling} shows an ablation study on the importance of up-sampling the lengths of MT inputs to match the lengths of discrete speech inputs (per \Sref{sec:mt}).
We found that $4$x up-sampling was best.
Note that without any up-sampling, ST/MT multi-tasking was actually slightly detrimental.
$6$x up-sampling was not the best even though speech and text have equal lengths on average, suggesting that the alignment of each text token to corresponding speech tokens is not uniform.

\section{Relation to Prior Work}
\label{sec:prior-work}

Now that we have presented our approach and results, we'll highlight the technical and empirical \textbf{novelty} of our work.

First, our work is closely related to AudioPalm \cite{rubenstein2023audiopalm}; both of our methods achieve hard ST/MT multi-tasking by discretizing speech.
On the surface this makes our methods look quite similar, but AudioPalm focuses on initializing speech models from Palm.
In fact, their results show that their models are deficient when trained from scratch (see Table 6 in their paper).
We take the exact opposite approach: we first confirm that our method improves training from scratch before adding external MT data and initialization from textual pre-trained models.
This is a major technical difference in itself, but it also allows us to investigate several empirical novelties.
Namely, we are able to show the individual effects of hard parameter sharing, external MT data, and transfer learning.
These three effects are conflated within the experimental setup of the AudioPalm paper which is more focused on demonstrating performance at scale.
All told, we view these works as complementary -- this work focuses on a set of sequence-to-sequence models commonly used in ST and other speech processing tasks (CTC, AED, CTC/Attn, RNN-T) which are distinct from AudioPalm's decoder-only model.

Second, our work follows a long line of prior works which investigate ST/MT multi-tasking \cite{ao2021speecht5, bapna2022mslam, tang2022unified, ye2022cross, fang-etal-2022-stemm, chen2022maestro, cheng2023m, wu2023decoder, ouyang-etal-2023-waco}.
The common theme amongst these approaches is soft parameter sharing, which is a major difference compared to our approach.
Further, we examine a larger set of Seq2Seq models to demonstrate general applicability.

\section{Conclusion}
We present a method for ST/MT multi-tasking with hard parameter sharing, which is not trivially achieved due to the speech-text modality gap.
Our approach resolves this by pre-processing speech into discrete sequences of tokens.
This allows us to build Seq2Seq models capable of ingesting speech and text via an input vocabulary consisting of discrete speech and text tokens.
Given the consistent improvements in ST, we will apply this approach to spoken language understanding and speech summarization in the future.
\blfootnote{Brian and Shinji are supported by the HLTCOE at JHU. This work used NCSA Delta (project CIS210014) from ACCESS through NSF grants \#2138259, \#2138286, \#2138307, \#2137603, and \#2138296.}

\vfill\pagebreak

\section{References}
{
\printbibliography
}

\end{document}